\documentclass[10pt,twocolumn,letterpaper]{article}

\usepackage{iccv}
\usepackage{times}
\usepackage{epsfig}
\usepackage{graphicx}
\usepackage{amsmath}
\usepackage{amssymb}

\usepackage[utf8]{inputenc} %
\usepackage[T1]{fontenc}    %
\usepackage{url}            %
\usepackage{booktabs}       %
\usepackage{amsfonts}       %
\usepackage{nicefrac}       %
\usepackage[shrink=10]{microtype} %
\usepackage{enumitem}

\usepackage{capt-of}

\usepackage{color}

\definecolor{turquoise}{cmyk}{0.65,0,0.1,0.3}
\definecolor{purple}{rgb}{0.65,0,0.65}
\definecolor{dark_green}{rgb}{0, 0.5, 0}
\definecolor{orange}{rgb}{0.8, 0.6, 0.2}
\definecolor{red}{rgb}{0.8, 0.2, 0.2}
\definecolor{blueish}{rgb}{0.0, 0.3, 1}
\definecolor{light_gray}{rgb}{0.7, 0.7, .7}
\definecolor{pink}{rgb}{1, 0, 1}
\definecolor{dark_red}{rgb}{0.5, 0, 0}

\newcommand{\citefig}[1]{Fig.~\ref{#1}}

\newcommand{\citetab}[1]{Table~\ref{#1}}
\newcommand{\citesec}[1]{Section~\ref{#1}}

\newcommand\para[1]{\smallskip\noindent\textbf{#1}}

\DeclareMathOperator*{\argmin}{argmin}

\usepackage[pagebackref=true,breaklinks=true,letterpaper=true,colorlinks,bookmarks=false]{hyperref}

\iccvfinalcopy %

\def\iccvPaperID{3034} %

\ificcvfinal\fi

\begin{document}

\title{Infinite Nature:\linebreak Perpetual View Generation of Natural Scenes from a Single Image}
\frenchspacing

\newcommand*\samethanks[1][\value{footnote}]{\footnotemark[#1]}
\newcommand\blfootnote[1]{%
  \begingroup
  \renewcommand\thefootnote{}\footnote{#1}%
  \endgroup
}

\author{
Andrew Liu\thanks{Equal Contribution} \hspace{0.5in} 
Richard Tucker\samethanks \hspace{0.5in}
Varun Jampani\\ 
Ameesh Makadia\hspace{0.5in}
Noah Snavely\hspace{0.5in}
Angjoo Kanazawa
\vspace{5pt}
\\
{Google Research} 
}

\twocolumn[{%
\renewcommand\twocolumn[1][]{#1}%
\maketitle
\ificcvfinal\thispagestyle{empty}\fi

\begin{center}
    \newcommand{\teaserwidth}{\textwidth}
    \vspace{-7mm}
    \centerline{
    \includegraphics[width=\teaserwidth]{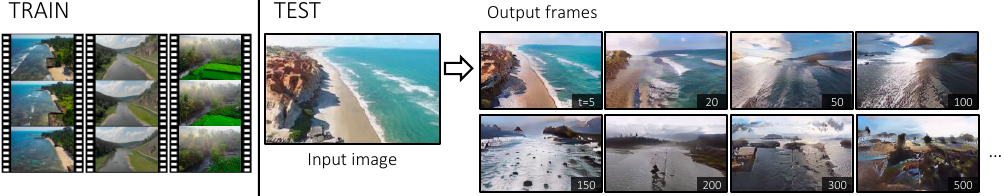}
     }
    \captionof{figure}{{\bf Perpetual View Generation.} Using a collection of aerial videos of nature scenes for training (left), our method learns to take a single image and perpetually generate novel views for a camera trajectory covering a long distance (right). %
    Our method can successfully generate hundreds of frames of an aerial video from a single input image (up to 500 shown here). %
    }
    \vspace{-1mm}
	\label{fig:teaser}

\end{center}%
}]

\begin{abstract}
\vspace{-2mm}
We introduce the problem of \textit{perpetual view generation}---long-range generation of novel views corresponding to an arbitrarily long camera trajectory given a single image. %
This is a challenging problem that goes far beyond the capabilities of %
current view synthesis methods, which
quickly degenerate when presented with large camera motions. 
Methods for video generation
also have limited ability to 
produce long sequences and are often agnostic to scene geometry.
We take a hybrid approach that integrates both geometry and image synthesis in an iterative `\emph{render}, \emph{refine} and \emph{repeat}' framework, allowing for long-range generation that cover large distances after hundreds of frames. Our approach can be trained from a set of monocular video sequences. 
We %
propose
a dataset of aerial footage of coastal scenes, and compare our method with recent view synthesis and conditional video generation baselines, showing that it can generate plausible scenes
for much longer time horizons over large camera trajectories compared to existing methods. 

\noindent Project page at
\href{https://infinite-nature.github.io/}{https://infinite-nature.github.io}.

\end{abstract}
\blfootnote{* indicates equal contribution}

\vspace{-6mm}
\section{Introduction}
Consider the input image of a coastline 
in \citefig{fig:teaser}.
Imagine flying through this scene as a bird. %
Initially, we would see objects grow in our field of view as we approach them. Beyond, we might find a wide ocean or new islands. At the shore, we might see cliffs or beaches, while inland there could be mountains or forests.
As humans, we are adept at imagining a plausible world from a single picture, based on our own experience.

How can we emulate this ability on a computer? One approach would be to attempt to generate an entire 3D planet with high-resolution detail from a single image. However, this would be extremely expensive and far beyond the current state of the art. So, we pose the more tractable problem of \textit{perpetual view generation}: given a single image of scene, the task is to synthesize a video corresponding to an arbitrary camera trajectory.
Solving this problem can have applications in content creation, novel photo interactions, and methods that use learned world models like model-based reinforcement learning.

Perpetual view generation, though simple to state, is an extremely challenging task. As the viewpoint moves, we must extrapolate new content in unseen regions and also synthesize new detail in existing regions that are now closer to the camera. Two active areas of research, video synthesis and view synthesis, both fail to scale to this problem for different reasons.

\begin{figure*}[t]
  \centering
  \includegraphics[width=\textwidth]{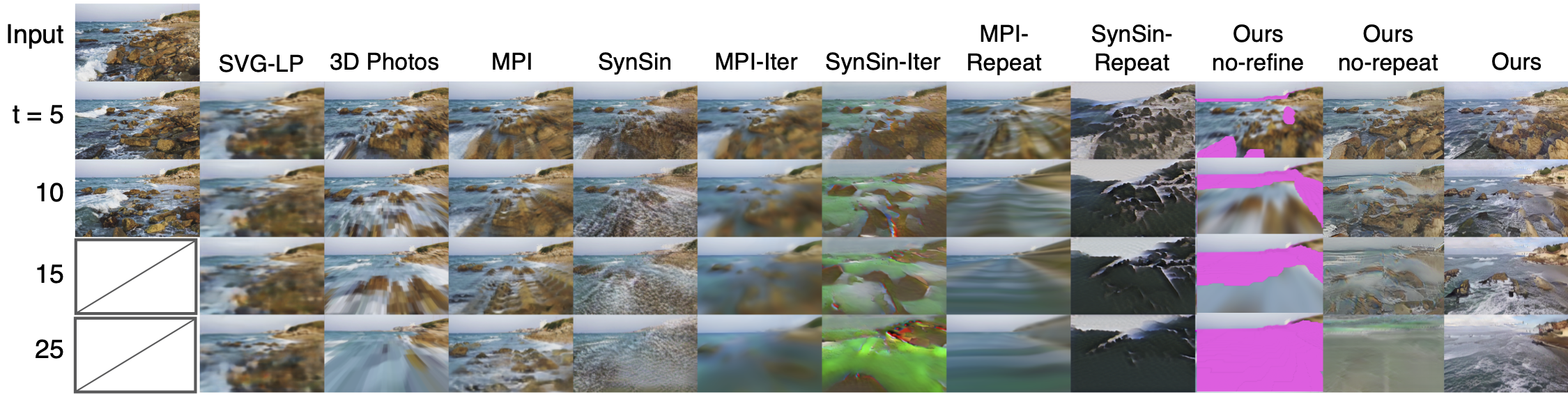}
  \vspace{-.8em}
  \caption{{\bf Qualitative comparison over time.} We show a generated sequence for each method at different time steps. Note that we only have ground truth images for 10 frames; the subsequent frames are generated using an extrapolated trajectory. Pink region in Ours no-refine indicate missing content uncovered by the moving camera.}
  \label{fig:main}
  \vspace{-.2em}
\end{figure*}

Recent video synthesis methods apply developments in image synthesis~\cite{karras2019style} to the temporal domain, or rely on recurrent models \cite{denton2018stochastic}. But they can generate only limited numbers of novel frames (e.g., 25 \cite{villegas2019high} or 48 frames \cite{DVD}).
Additionally, such methods often neglect an important element of the video's structure---they model neither scene geometry nor camera movement.
In contrast, many view synthesis methods do take advantage of geometry to synthesize high-quality novel views. However, these approaches can only operate within a limited range of camera motion. As shown in Figure \ref{fig:main}, once the camera moves outside this range, such methods fail catastrophically.

We propose a hybrid framework that takes advantage of both geometry and image synthesis techniques to address these challenges. %
We use disparity maps to represent a scene's geometry, %
and decompose the perpetual view generation task into the framework of \emph{render}-\emph{refine}-and-\emph{repeat}. First, we \textbf{render} the current frame from a new viewpoint, using disparity to ensure that scene content 
moves in a geo\-metrically correct manner. Then, we \textbf{refine} the resulting image and geometry. This step adds detail and synthesizes new content in areas that require inpainting or outpainting.
Because we refine both the image and disparity, the whole process can be \textbf{repeated} in an recurrent manner, allowing for perpetual generation with arbitrary trajectories.

To train our system, we curated a large dataset of drone footage of nature and coastal scenes from over 700 videos, spanning 2 million frames. 
We run a structure from motion pipeline to recover 3D camera trajectories, and refer to this as the Aerial
Coastline Imagery Dataset (ACID).
Our trained model can generate sequences of hundreds of frames while maintaining the aesthetic feel of an aerial coastal video, even though after just a few frames, the camera has moved beyond the limits of the scene depicted in the initial view.

Our experiments show that our novel render-refine-repeat framework, with propagation of geometry via disparity maps, is key to tackling this problem. 
Compared to recent view synthesis and video generation baselines, our approach can produce plausible frames for much longer time horizons. 
This work represents a significant step towards perpetual view generation, though it has limitations such as a lack of global consistency in the hallucinated world. We believe our method and dataset will lead to further advances in generative methods for large-scale scenes.

\begin{figure*}[t!]
  \centering
  \includegraphics[width=\textwidth]{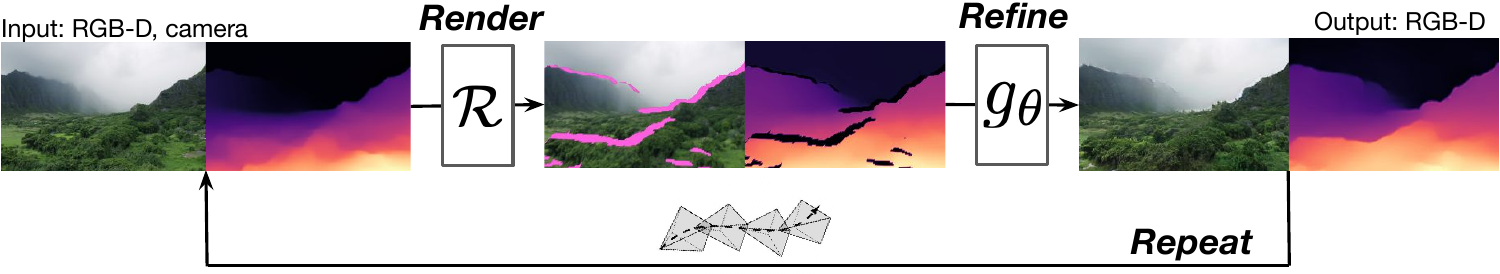}
  \caption{{\bf Overview.} We first \emph{render} an input image to a new camera view using the disparity.
  We then \emph{refine} the image, synthesizing and super-resolving missing content. As we output both RGB and geometry, this process can be \emph{repeated} for perpetual view generation.}
  \label{fig:overview}
\end{figure*}
\section{Related Work}
\noindent\textbf{Image extrapolation.}
Our work is inspired by the seminal work of Kaneva \etal \cite{Kaneva_2010}, which proposed a non-parametric %
approach for generating `infinite' images through stitching 2D-transformed images, and by patch-based non-parametric approaches for image extension \cite{schodl2000video,Barnes:2009:PAR}. %
We revisit the `infinite images' concept in a learning framework that also reasons about the 3D geometry behind each image. 
Also related to our work are recent deep learning approaches to the problem of \emph{outpainting}, i.e., inferring unseen content outside image boundaries \cite{wang2019wide,yang2019very,teterwak2019boundless}, as well as 
\emph{inpainting}, the task of filling in 
missing content within an image \cite{hays2007scene,yu2018free}. These approaches use adversarial frameworks and semantic information for in/outpainting. %
Our problem also incorporates aspects of super-resolution \cite{glasner2009super,ledig:2017:photo}. Image-specific GAN methods also demonstrate a form of image extrapolation and super-resolution of textures and natural images \cite{zhou2018non,shocher2018zero,shaham2019singan,shocher2018ingan}.
In contrast to the above methods, we reason about the 3D geometry behind each image and study image extrapolation in the context of temporal image sequence generation.%

\para{View synthesis.}
Many view synthesis methods operate by interpolating between multiple views of a scene~\cite{levoy:1996:lightfield,chaurasia:2013:depth,mildenhall:2019:llff,flynn:2019:deepview,extremeview}, although recent work can generate new views from just a single input image, as in our work~\cite{chen2019mono,tulsiani:2018:lsi,niklaus:2019:kenburns,single_view_mpi,shi:2014:lightfield,chen2019monocular}. 
However, in both %
settings, most methods only allow for a very limited range of output viewpoints. Even methods that explicitly allow for view extrapolation (not just interpolation) typically restrict the camera motion to 
small regions around a reference view~\cite{zhou:2018:stereo,srinivasan:2019:boundaries,choi2019extreme}.

One factor that limits camera motion is that many methods construct a static
scene representation, such as a layered depth image \cite{tulsiani:2018:lsi,Shih3DP20}, multiplane image \cite{zhou:2018:stereo,single_view_mpi}, point cloud \cite{niklaus:2019:kenburns,wiles2020synsin}, or radiance field \cite{yu2020pixelnerf,grf2020}, and 
inpaint disoccluded regions. Such representations can allow for fast rendering, but the range of viable camera positions is limited by the finite bounds of the scene representation.
Some methods augment this scene representation paradigm, enabling a limited increase in the range of output views. %
Niklaus \etal\ perform inpainting \textit{after} rendering \cite{niklaus:2019:kenburns}, while
SynSin uses a post-rendering refinement network to produce realistic images from feature point-clouds \cite{wiles2020synsin}. We take inspiration from these methods by rendering and then refining our output. In contrast, however, our system does not construct a single 3D representation of a scene. Instead we proceed iteratively, generating each output view from the previous one, and producing a geometric scene representation in the form of a disparity map for each frame. 

Some methods use video as training data.
Monocular depth can be learned from 3D movie left-right camera pairs \cite{ranftl:2020:towards} or from video sequences analysed with structure-from-motion techniques \cite{chen:2019:youtube3d}. Video can also be directly used for view synthesis \cite{single_view_mpi,wiles2020synsin}. These methods use pairs of images, whereas our model is trained on sequences of several widely-spaced frames since we want to generate long-range video.

\para{Video synthesis.}
Our work is related to methods that generate a video sequence from one or more images \cite{vondrick2016generating,finn2016unsupervised,vondrick2017generating,denton2018stochastic,tulyakov2018mocogan,Ye_2019_ICCV}. Many such approaches have focused on predicting the future of dynamic objects with a static camera, often using simple videos of humans walking \cite{blank2005actions} or robot arms \cite{finn2016unsupervised}. 
In contrast, we focus on mostly static scenes with a moving camera, using real aerial videos of nature.
Some recent research addresses video synthesis from in-the-wild videos with moving cameras \cite{DVD,villegas2019high}, but without taking geometry explicitly into account, and with strict limits on the the length of the generated video.
In contrast, in our work the movement of pixels from camera motion is explicitly modeled using 3D geometry.
\begin{figure*}[t]
  \centering
  \includegraphics[width=\textwidth]{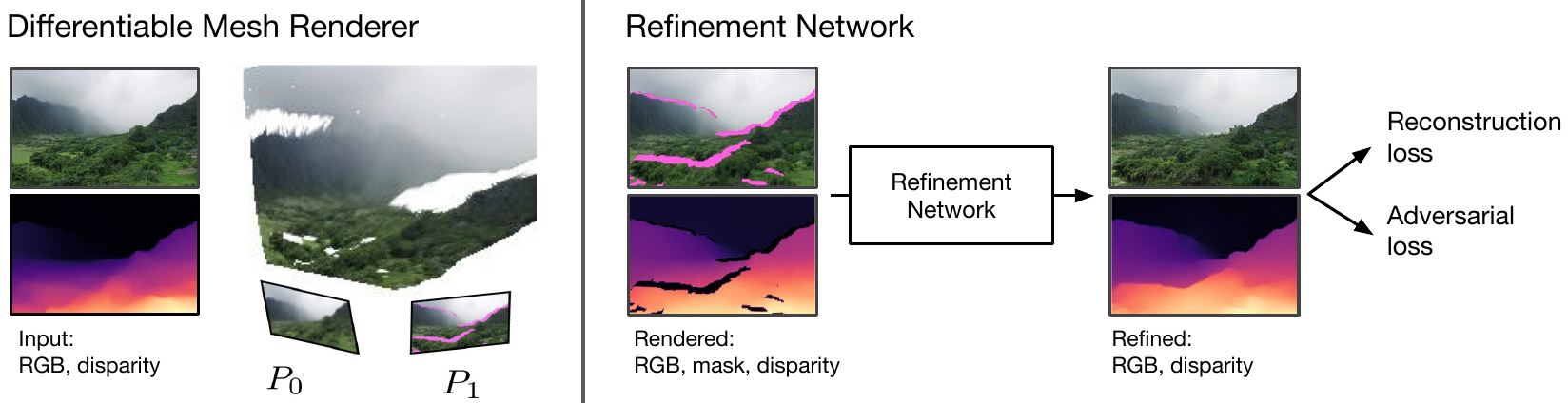}
  \caption{{\bf Illustration of the rendering and refinement steps.} Left: Our differentiable rendering stage takes a paired RGB image and disparity map from viewpoint $P_0$ and creates a textured mesh representation, which we render from a new viewpoint $P_1$, warping the textures, adjusting disparities, and returning a binary mask representing regions to fill in. Right: The refinement stage takes the output of the renderer and uses a deep network to fill in holes and add details. The output is a new RGB image and disparity map that can be supervised with reconstruction and adversarial losses.}
  \label{fig:arch}
  \vspace{-.5em}
\end{figure*}

\section{Perpetual View Generation}
Given an RGB image $I_0$ and a camera trajectory $(P_0, P_1, P_2, \dots)$ of arbitrary length, our task is to output a new image sequence $(I_0, I_1, I_2, \dots)$ that forms a video depicting a flythrough of the scene captured by the initial view. The trajectory is a series of 3D camera poses $P_t = \left(\begin{smallmatrix}
R^{3\times 3} & \mathbf{t}^{3\times 1} \\
0 & 1\end{smallmatrix}\right),$ where $R$ and $\mathbf{t}$ are 3D rotations and translations, respectively. In addition, each camera has an intrinsic matrix $K$. At training time camera data is obtained from video clips via structure-from-motion as in \cite{zhou:2018:stereo}. At test time, the camera trajectory may be pre-specified, generated by an auto-pilot algorithm, or controlled via a user interface.

\subsection{Approach: Render, Refine, Repeat}
Our framework applies established techniques (3D rendering, image-to-image translation, auto-regressive training) in a novel combination.
We decompose \emph{perpetual view generation} into the three steps, as illustrated in Figure~\ref{fig:overview}:
\begin{enumerate}[nosep]
\item\textbf{Render} a new view from an old view, by warping 
the image according to a disparity map using a differentiable renderer,
\item\textbf{Refine} the rendered view and geometry to fill in missing content and add detail where necessary,
\item\textbf{Repeat} this process, propagating \textit{both image and disparity} to generate each new view from the one before.
\end{enumerate}
Our approach has several desirable characteristics. \textit{Representing geometry} with a disparity map allows much of the heavy lifting of moving pixels from one frame to the next to be handled by differentiable rendering, %
ensuring local temporal consistency. %
The synthesis task then becomes one of \textit{image refinement}, which comprises: 1) inpainting disoccluded regions 2) outpainting of new image regions and 3) super-resolving image content.
Because every step is \textit{fully differentiable}, we can train our refinement network by backpropagating through several view synthesis iterations. Our \textit{auto-regressive} framework means that novel views may be infinitely generated with explicit view control, even though training data is finite in length.

Formally, for an image $I_t$ with pose $P_t$ we have an associated disparity (i.e., inverse depth) map $D_t \in \mathbb{R}^{H \times W}$, and we compute the next frame $I_{t+1}$ and its disparity $D_{t+1}$ as
\begin{align}
  \hat{I}_{t+1}, \hat{D}_{t+1}, \hat{M}_{t+1}
  &= \mathcal{R}(I_{t}, D_t, P_t, P_{t+1}), \\
  I_{t+1}, D_{t+1} &= g_\theta(\hat{I}_{t+1}, \hat{D}_{t+1}, \hat{M}_{t+1}).
\end{align}
Here, $\hat{I}_{t+1}$ and $\hat{D}_{t+1}$ are the result of rendering 
the image $I_t$ and disparity $D_t$ from the new camera $P_{t+1}$, using a differentiable renderer $\mathcal{R}$~\cite{Genova_2018_CVPR}. This function also returns a mask $\hat{M}_{t+1}$ indicating which regions of the image are missing and need to be filled in. The refinement network $g_\theta$ then inpaints, outpaints and super-resolves these inputs to produce the next frame $I_{t+1}$ and its disparity $D_{t+1}$. The process is repeated iteratively for $T$ steps during training, and at test time for an arbitrarily long camera trajectory. Next we discuss each step in detail.

\para{Geometry and Rendering.}
Our render step $\mathcal{R}$ uses a differentiable mesh renderer~\cite{Genova_2018_CVPR}. First, we convert each pixel coordinate $(u, v)$ in $I_t$ and its corresponding disparity $d$ in $D_t$ into a 3D point in the camera coordinate system: $(x,y,z) = K^{-1}(u, v, 1)/d$. We then convert the image into a 3D triangular mesh where each pixel is treated as a vertex connected to its neighbors, ready for rendering.

To avoid stretched triangle artifacts at depth discontinuities and aid our refinement network by identifying regions to be inpainted, we compute a per-pixel binary mask $M_t\in \mathbb{R}^{H\times W}$ by thresholding the gradient of the disparity image $\nabla \hat{D_t}$, computed with a a Sobel filter:
\begin{equation}
    M_t = \begin{cases} 0 & \text{where } ||\nabla \hat{D_t}|| > \alpha, \\
    1 & \text{otherwise.}
    \end{cases}
\end{equation}
We use the 3D mesh to render both image and mask from the new view $P_{t+1}$, and multiply the rendered image element-wise by the rendered mask to give $\hat{I}_{t+1}$.
The renderer also outputs a depth map as seen from the new camera, which we invert and multiply by the rendered mask to obtain $\hat{D}_{t+1}$.
This use of the mask ensures that any regions in $\hat{I}_{t+1}$ and $\hat{D}_{t+1}$ that were occluded in $I_t$ are masked out and set to zero (along with regions that were outside the field of view of the previous camera). These areas are ones that the refinement step will have to inpaint (or outpaint). See Figures \ref{fig:overview} and \ref{fig:arch} for examples of missing regions shown in pink.

\para{Refinement and Synthesis.}
Given the rendered image $\hat{I}_{t+1}$, its disparity $\hat{D}_{t+1}$ and its mask $\hat{M}_{t+1}$, our next task is to refine this image, which includes blurry regions and missing pixels. In contrast to prior inpainting work~\cite{yu2018generative,teterwak2019boundless}, the refinement network also has to perform super-resolution and thus we cannot use a compositing operation in refining the rendered image. 
Instead we view the refinement step as a generative image-to-image translation task, and adopt the state-of-the-art SPADE network architecture~\cite{park2019SPADE} for our 
$g_\theta$, which directly outputs $I_{t+1}, D_{t+1}$. We encode $I_0$ to provide the additional GAN noise input required by this architecture. See the appendix for more details.

\para{Rinse and Repeat.}
The previous steps allow us to generate a single novel view. A crucial aspect of our approach is that we refine not only RGB but also disparity, so that scene geometry is propagated between frames. With this setup, we can use the refined image and disparity as the next input to train in an auto-regressive manner, with losses backpropagated over multiple steps. 
Other view synthesis methods, although not designed in this manner, may also be trained and evaluated in a recurrent setting, although naively repeating these methods without propagating the geometry as we do requires the geometry to be re-inferred from scratch in every step. As we show in \citesec{sec:evaluation}, training and evaluating these baselines with a \textit{repeat} step is still insufficient for perpetual view generation.

\para{Geometric Grounding to Prevent Drift.}
\label{sec:grounding}
A key challenge in generating long sequences is dealing with the accumulation of errors \cite{ross2011reduction}. In a system where the current prediction affects future outputs, small errors in each iteration can compound, eventually generating predictions outside the distribution seen during training and causing unexpected behaviors. %
Repeating the generation loop in the training process and feeding the network with its own output ameliorates drift and improves visual quality as shown in our ablation study (\citesec{sec:ablations}). However, we notice that the disparity in particular can still drift at test time, especially over time horizons far longer than seen during training. Therefore we add an explicit geometric re-grounding of the disparity maps.

Specifically, we take advantage of the fact that the rendering process provides the correct range of disparity from a new viewpoint $\hat{D}_{t+1}$ for visible regions of the previous frame.
The refinement network may modify these values as it refines the holes and blurry regions, which can lead to drift as the overall disparity becomes gradually larger or smaller than expected. However, we can geometrically correct this by rescaling the refined disparity map to the correct range by computing a scale factor $\gamma$ via solving
\begin{equation}
    \min_{\gamma} || M \odot (\log(\gamma D_{t+1}) - \log(\hat{D}_{t+1}))   ||
\end{equation}
By scaling the refined disparity by $\gamma$, our approach ensures that the disparity map stays at a consistent scale, which significantly reduces drift at test time as shown in \citesec{sec:perpetual}.

\begin{figure}
\includegraphics[width=\linewidth]{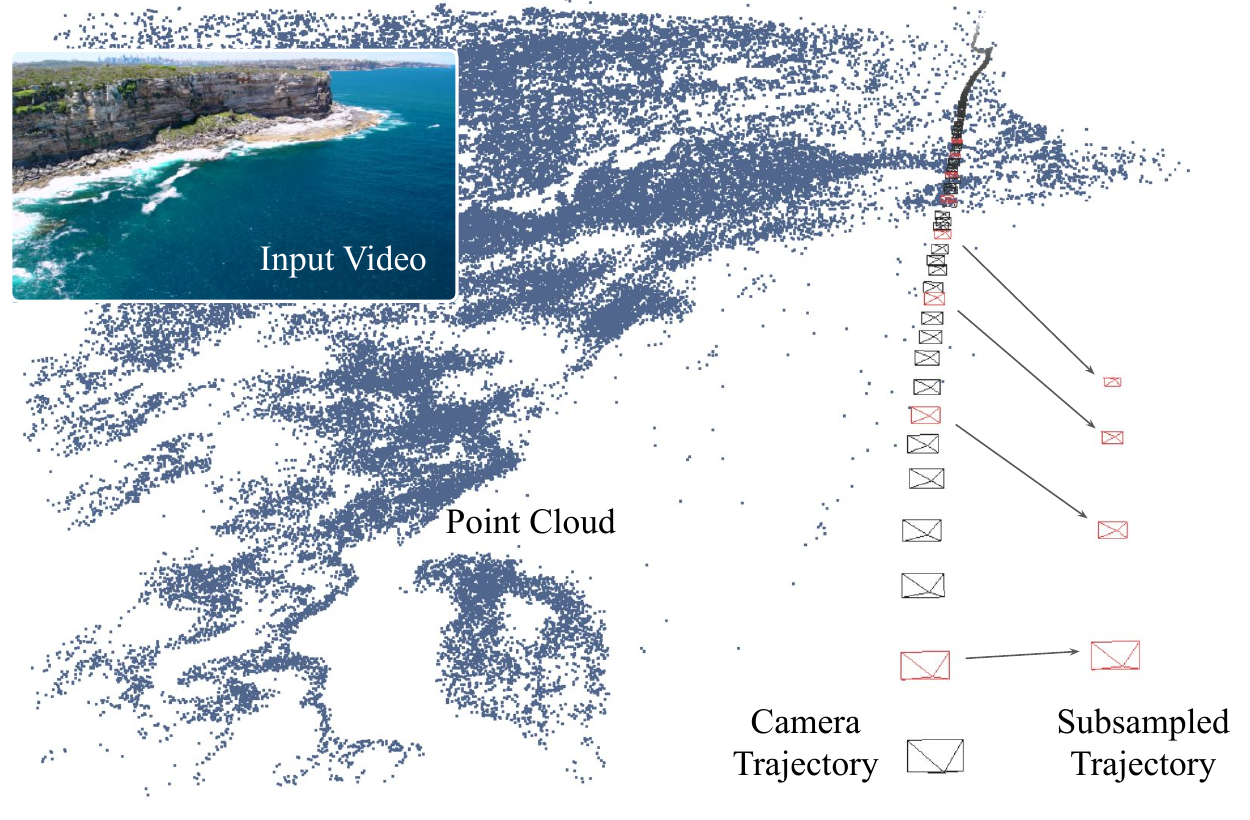}
\caption{\textbf{Processing video for ACID.} We run structure from motion on coastline drone footage collected from YouTube to create the Aerial Coastline Imagery Dataset (ACID). See \citesec{sec:data}.}
\vspace{-.5em}
\label{fig:dataset}
\end{figure}

\section{Aerial Coastline Imagery Dataset}
\label{sec:data}
Learning to generate long sequences requires real image sequences for training. Many existing datasets for view synthesis do not use sequences, but only a set of views from slightly different camera positions. Those that do have sequences are limited in length: RealEstate10K, for example, has primarily indoor scenes with limited camera movement \cite{zhou:2018:stereo}. To obtain long sequences with a moving camera and few dynamic objects, we turn to aerial footage of beautiful nature scenes available on the Internet. Nature scenes are a good starting point for our challenging problem, as GANs have shown promising results on nature textures \cite{shaham2019singan,shocher2018ingan}.
We collected 891 videos using keywords such as `coastal' and `aerial footage', and processed these videos with SLAM and structure from motion following the approach of Zhou \etal\ \cite{zhou:2018:stereo}, yielding over 13,000 sequences with a total of 2.1 million frames. We have released the list of videos and SfM camera trajectories. See \citefig{fig:dataset} for an illustrative example of our SfM pipeline running on a coastline video.

To obtain disparity maps for every frame, we use the off-the-shelf MiDaS single-view depth prediction method~\cite{ranftl:2020:towards}. We find that MiDaS is quite robust and produces sufficiently accurate disparity maps for our method. Because MiDaS disparity is only predicted up to scale and shift, it must first be rescaled to match our data. To achieve this, we use the sparse point-cloud computed for each scene during structure from motion. For each frame we consider only the points that were tracked in that frame, and use least-squares to compute the scale and shift that minimize the disparity error on these points. We apply this scale and shift to the MiDaS output to obtain disparity maps $(D_i)$ that are scale-consistent with the SfM camera trajectories $(P_i)$ for each sequence.

Due to the difference in camera motions between videos, we strategically sub-sample frames to ensure consistent camera speed in training sequences. See more details in the appendix.

\section{Experimental Setup}
\label{sec:setup}

\noindent \textbf{Losses.}
We train our approach on a collection of image sequences $\{I_t\}_{t=0}^T$
with corresponding camera poses $\{P_t\}_{t=0}^T$ %
and disparity maps for each frame $\{D_t\}_{t=0}^T$.
Following the literature on conditional generative models, we use an L1 reconstruction loss on RGB and disparity, a VGG perceptual loss on RGB \cite{johnson2016perceptual} and a hinge-based adversarial loss with a discriminator (and feature matching loss) \cite{park2019SPADE} for the $T$ frames that we synthesize during training. We also use a KL-divergence loss \cite{kingma2013auto} on our initial image encoder $\mathcal{L}_{\text{KLD}} = \mathcal{D}_{\text{KL}}(q(z|x) || \mathcal{N}(0, 1))$. Our complete loss function is
\begin{equation}
    \mathcal{L} = \mathcal{L}_{\text{reconst}} + \mathcal{L}_{\text{perceptual}} + \mathcal{L}_{\text{adv}} +  \mathcal{L}_{\text{feat matching}} + \mathcal{L}_{\text{KLD}}
\end{equation}
The loss is computed over all iterations and over all samples in the mini-batch.

\begin{figure*}[t]
  \begin{minipage}{.38\textwidth}
    \centering
   
    \resizebox{\textwidth}{!}{

\begin{tabular}{lcrr@{}p{1em}@{}r@{}c@{}}
\toprule
&& \multicolumn{2}{c}{Over frames 1--10}
&& Over frames&\ 1--50 \\
\cmidrule{3-4} \cmidrule{6-7}
Method    && $\textrm{LPIPS}\downarrow$ & $\textrm{MSE}\downarrow$ && $\textrm{FID}\downarrow$    \\
\midrule
\multicolumn{6}{l}{\hspace{-.5em}\textit{Baseline methods}} \\
SVG-LP \cite{denton2018stochastic}    && 0.60 & 0.020 && 135.9    \\
SynSin \cite{wiles2020synsin}    && 0.32 & \textbf{0.018} && 98.1  \\
MPI \cite{single_view_mpi} && 0.35 & 0.019 && 65.0      \\
3D Photos \cite{Shih3DP20} && \textbf{0.30} & 0.020 && 123.6     \\
\midrule
\multicolumn{6}{l}{\hspace{-.5em}\textit{Applied iteratively at test time}} \\
SynSin--Iterated    && 0.40 & 0.021 && 143.6  \\
MPI--Iterated   && 0.47 & 0.020 && 201.2      \\
\midrule
\multicolumn{6}{l}{\hspace{-.5em}\textit{Trained with repeat ($T=5$)}} \\
SynSin--Repeat && 0.44 & 0.036 && 153.3 \\
MPI--Repeat && 0.55 & 0.020 && 203.0 \\
\midrule
Ours      && 0.32 & 0.020 && \textbf{50.6} \\
\bottomrule
\end{tabular}

}
\vspace{.5em}
\captionof{table}{{\bf Quantitative evaluation.}
We compute LPIPS and MSE against ten frames of ground truth, and FID-50 over 50 frames generated from an input test image. See \citesec{sec:short_to_medium}.}
\label{tab:comp}
\end{minipage}%
\hfill
\begin{minipage}{.57\textwidth}
\centering\includegraphics[width=1.0\linewidth]{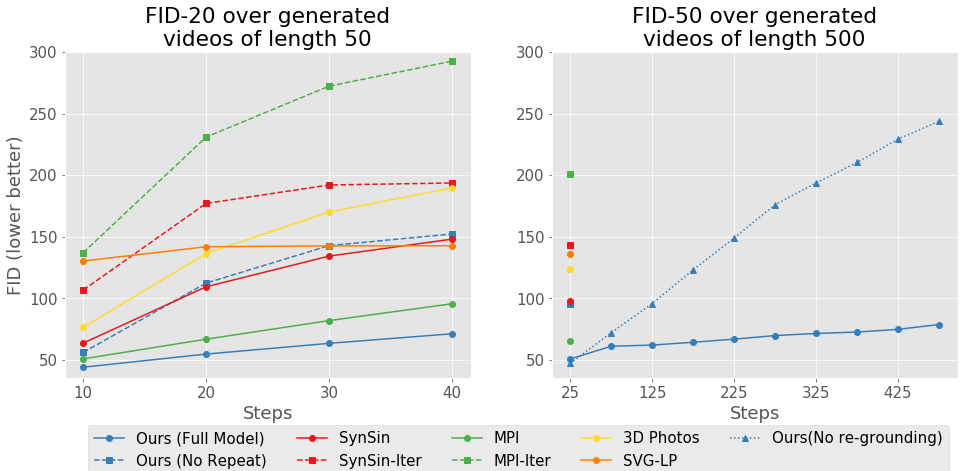}
\caption{\textbf{FID over time.} Left: FID-20 over time for 50 frames generated by each method. Right: FID-50 over \emph{500} frames generated by our method via autopilot. For comparison, we plot FID-50 for the baselines on the first 50 steps. Despite generating sequences an order of magnitude longer, our FID-50 is still lower than that of the baselines. See Sec.~\ref{sec:short_to_medium}, \ref{sec:perpetual}.}
\label{fig:plot}
\end{minipage}
\vspace{-1em}
\end{figure*}

\para{Metrics.} Evaluating the quality of the generated images in a way that correlates with human judgement is a challenge. We use the Fr\'echet inception distance (FID), a common metric used in evaluating generative models of images. FID computes the difference between the mean and covariance of the embedding of real and fake images through a pretrained Inception network \cite{fid} to measure the realism of the generated images as well as their diversity. %
We precompute real statistics using 20k real image samples from our dataset.
To measure changes in generated quality over time, we report FID over a sliding window: we write FID-$w$ at $t$ to indicate a FID value computed over all image outputs within a window of width $w$ centered at time $t$, i.e.\ $\{I_i\}$ for $t-w/2 < i \leq t+w/2$.
For short-range trajectories where ground truth images are available, we also report mean squared error (MSE) and LPIPS~\cite{zhang:2018:lpips}, a perceptual similarity metric that correlates better with human perceptual judgments than traditional metrics such as PSNR and SSIM. 

\para{Implementation Details.} We train our model with $T=5$ steps of render-refine-repeat at an image resolution of 160 $\times$ 256 (as most aerial videos have a 16:9 aspect ratio). The choice of $T$ is limited by both memory and available training sequence lengths. The refinement network architecture is the same as that of SPADE generator in \cite{park2019SPADE}, and we also employ the same multi-scale discriminator. We implement our models in TensorFlow, and train with a batch size of 4 over 10 GPUs for 7M iterations, which takes about 8 days. We then identify the model checkpoint with the best FID score over a validation set.
\section{Evaluation} 
\label{sec:evaluation}

We compare our approach with three recent state-of-the-art single-image view synthesis methods---the 3D Photography method (henceforward `3D Photos') \cite{Shih3DP20}, SynSin \cite{wiles2020synsin}, and single-view MPIs \cite{single_view_mpi}---as well as the SVG-LP video synthesis method \cite{denton2018stochastic}. We retrain each method on our ACID training data, with the exception of 3D Photos which is trained on in-the-wild imagery and, like our method, takes MiDaS disparity as an input. SynSin and single-view MPI 
were trained at a resolution of $256\times256$. SVG-LP takes two input frames for context, and operates at a lower resolution of $128\times128$.

The view synthesis baseline methods were not designed for long camera trajectories; every new frame they generate comes from the initial frame $I_0$ even though after enough camera movement there may be very little overlap between the two. Therefore we also compare against two variants of each of these methods. First, variants with \textit{iterated evaluation} (Synsin--Iterated, MPI--Iterated): these methods use the same trained models as their baseline counterparts, but we apply them iteratively at test time to generate each new frame from the \textit{previous} frame rather than the initial one.
Second, variants \textit{trained with repeat} (Synsin--Repeat,  MPI--Repeat): these methods are trained autoregressively, with losses backpropagated across $T=5$ steps, as in our full model.
(We omit these variations for the 3D Photos method, which was unfortunately too slow to allow us to apply it iteratively, and which we are not able to retrain.)

\begin{figure*}[h]
  \centering
  \includegraphics[width=\textwidth]{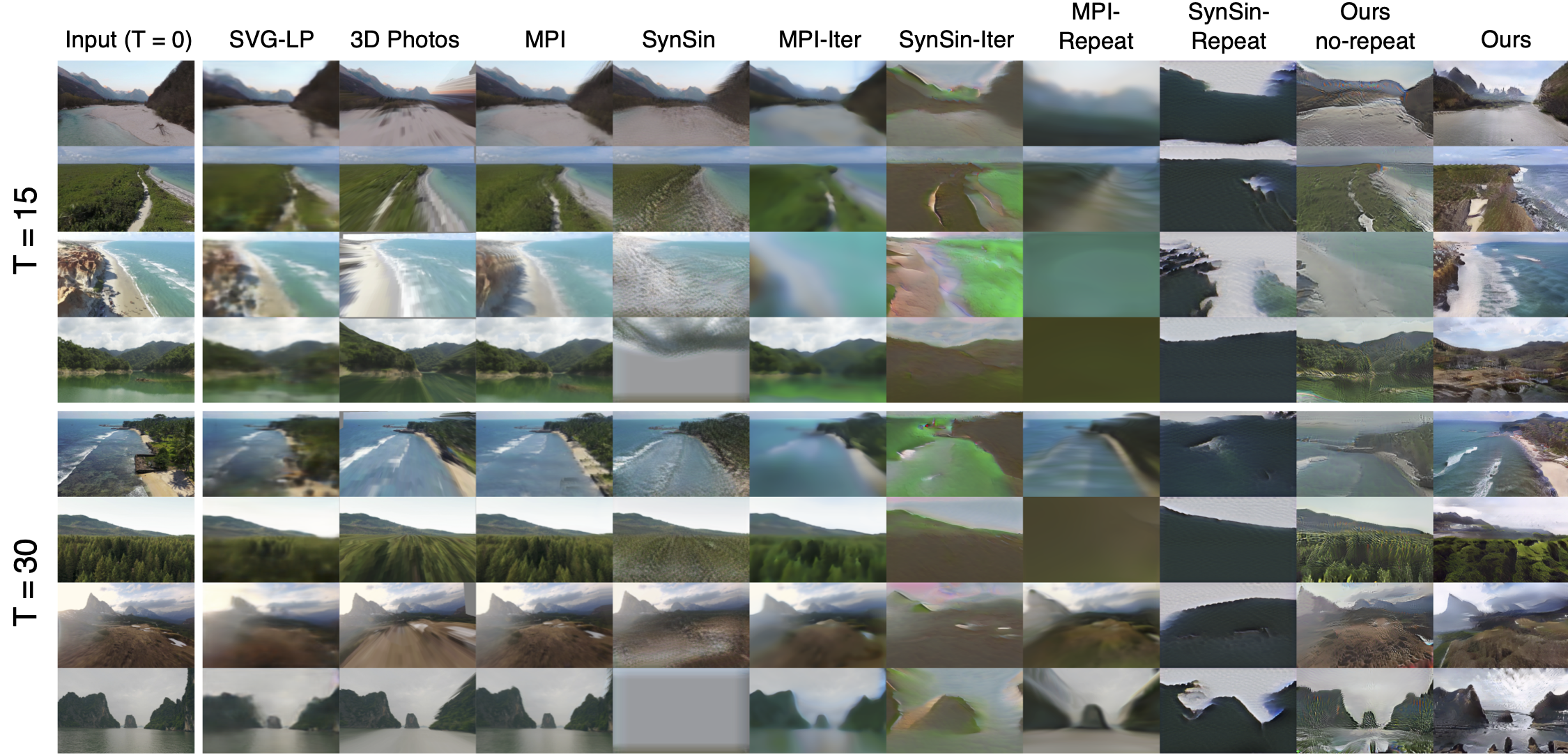}
  \vspace{-.5em}
  \caption{ {\bf Qualitative comparison.} We show the diversity and quality of many generated results for each method on the t=15 and 30 frame generation. Competing approaches result in missing or unrealistic frames. Our approach is able to generate plausible views of the scene.}
  \label{fig:sample}
\end{figure*}
\begin{figure*}[h]
  \centering
  \includegraphics[width=\textwidth]{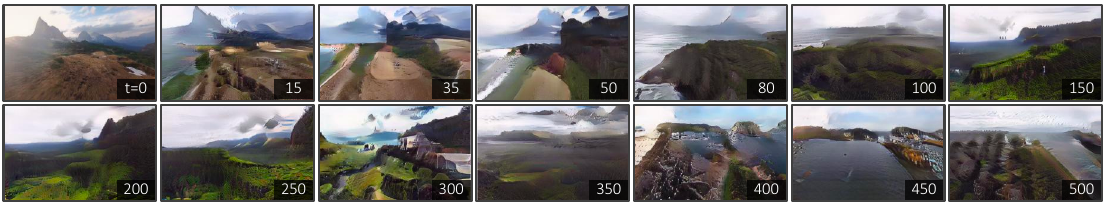}
  \vspace{-.8em}
  \caption{{\bf Long trajectory generation}. From a single image, our approach can generate 500 frames of video without suffering visually. Please see the supplementary video for the full effect. }
  \label{fig:longsequence}
  \vspace{-.2em}
\end{figure*}

\subsection{Short-to-medium range view synthesis}
\label{sec:short_to_medium}
To evaluate short-to-medium-range synthesis, we select ACID test sequences with an input frame and 10 subsequent ground truth frames (subsampling as described in the appendix), with the camera moving forwards at an angle of up to 45$^\circ$. Although our method is trained on all types of camera motions, this forward motion is appropriate for comparison with view synthesis methods which are not designed to handle extreme camera movements.

We then extrapolate the camera motion from the last two frames of each sequence to extend the trajectory for an additional 40 frames. To avoid the camera colliding with the scene, we check the final camera position against the disparity map of the last ground-truth frame, and discard sequences in which it is outside the image or at a depth large enough to be occluded by the scene.

This yields a set of 279 sequences with camera trajectories of 50 steps and ground truth images for the first 10 steps. For short-range evaluation, we compare to ground truth on the first 10 steps. For medium-range evaluation, we compute FID scores over all 50 frames.

We apply each method to these sequences to generate novel views corresponding to the camera poses in each sequence (SVG-LP is the exception in that it does not take account of camera pose). See results in \citetab{tab:comp}.
While our goal is perpetual view generation, we find that our approach is competitive with recent view synthesis approaches for short-range synthesis on LPIPS and MSE metrics. For mid-range evaluation, we report FID-$50$ over 50 generated frames. 
Our approach has a dramatically lower FID-50 score than other methods, reflecting the more naturalistic look of its output. To quantify the degradation of each method over time, we report a sliding window FID-$20$ computed from $t=10$ to 40.
As shown in \citefig{fig:plot} (left), the image quality (measured by FID-$20$) of the baseline methods deteriorates more rapidly with increasing $t$ compared to our approach.

Qualitative comparisons of these methods are shown in \citefig{fig:main} and our supplementary video, which illustrates how the quality of each method's output changes over time.
Notable here are SVG-LP's blurriness and inability to predict any camera motion at all; the increasingly stretched textures of 3D Photos' output; and the way the MPI-based method's individual layers become noticeable. SynSin does the best job of generating plausible texture, but still produces holes after a while and does not add new detail. 

The --Iterated and --Repeat variants are consistently worse than the original SynSin and MPI methods, which suggests that simply applying an existing method iteratively, or retraining it autoregressively, is insufficient to deal with large camera movement. These variants show more drifting artifacts than their original versions, likely because (unlike our method), they do not propagate geometry from step to step. The MPI methods additionally become very blurry on repeated application, as they have no ability to add detail, lacking our refinement step.

In summary, our thoughtful combination of render-refine-repeat shows better results than these existing methods and
variations. Figure \ref{fig:sample} shows additional qualitative results from generating 15 and 30 frames using
on a variety of inputs.

\subsection{Ablations}
\label{sec:ablations}
We investigate the benefit of training over multiple iterations of our \textit{render-refine-repeat} loop by also training our model with $T=1$ (`No repeat'). As shown in in \citetab{tab:ablation}, the performance on short-range generation, as measured in LPIPS and MSE, is similar to our full model, but when we look at FID, we observe that this method generates lower quality images and that they get substantially worse with increasing $t$ (see \citefig{fig:plot}, left). This shows the importance of using a recurrent training setup to our method.

We next consider the \textit{refine} step. Omitting this step completely results in a larger and larger portion of the image being completely missing as $t$ increases: examples are shown as \textit{`Ours (no refine)'} in \citefig{fig:main}, where for clarity the missing pixels are highlighted in pink. In our full model, these regions are inpainted or outpainted by the refinement network at each step. Note also that even non-masked areas of the image are much blurrier when the refinement step is omitted, showing the benefit of the refinement network in super-resolving image content.

\citetab{tab:ablation} also shows results on two further variations of our refinement step. First, replacing our refinement network with a simpler U-Net architecture yields substantially worse results (`U-Net refinement'). Second, disabling geometric grounding (\citesec{sec:grounding}) also leads to slightly lower quality on this short-to-medium range view synthesis task (`No re-grounding').

\begin{table}
\begin{center}
\vspace{.5em}
\begin{tabular}{lrrr}
\toprule
Ablations    & LPIPS $\downarrow$& MSE    $\downarrow$        & \multicolumn{1}{l}{FID-50} $\downarrow$     \\
\midrule
Full Model    & 0.32 & \textbf{0.020} & \textbf{50.6} \\
No repeat ($T=1$) & \textbf{0.30} & 0.022 & 95.4     \\
U-net Refinement & 0.54 & 0.052 & 183.0 \\
No re-grounding & 0.34 & 0.022 & 64.3 \\
\bottomrule
\end{tabular}
\vspace{.8em}
\captionof{table}{\textbf{Ablations.} We ablate aspects of our model to understand their contribution to the overall performance. See \citesec{sec:ablations}.}
\label{tab:ablation}
\end{center}
\end{table}

\subsection{Perpetual view generation}
\label{sec:perpetual}
We also evaluate the ability of our model to perform perpetual view generation by synthesizing videos of 500 frames, using an \textit{auto-pilot} algorithm to create an online camera trajectory that avoids flying directly into the ground, sky or obstacles such as mountains.
This algorithm works iteratively in tandem with image generation to control the camera based on heuristics which measure the proportion of sky and of foreground obstacles in the scene. See the appendix for details.

We note that this task is exceptionally challenging and completely outside the capabilities of current generative \textbf{and} view synthesis methods. To further frame the difficulty, our refinement network has only seen videos of length 5 during training, yet we generate 500 frames for each of our test sequences. As shown in \citefig{fig:plot} (right), our FID-50 score over generated frames is remarkably robust: even after 500 frames, the FID is lower than that of all the baseline methods over 50 frames.
\citefig{fig:plot} also shows the benefit of our proposed geometric grounding: when it is omitted, the image quality gradually deteriorates, indicating that resolving drift is an important contribution.
 
\citefig{fig:longsequence} shows a qualitative example of long sequence generation. In spite of the intrinsic difficulty of generating frames over large distances, our approach retains something of the aesthetic look of coastline, generating new islands, rocks, beaches, and waves as it flies through the world. The auto-pilot algorithm can receive additional inputs (such as a user-specified trajectory or random elements), allowing us to generate diverse videos from a single image.
Please see the supplementary video for more examples and the full effect of these generated fly-through videos.

\subsection{User-controlled video generation}

Because our rendering step takes camera poses as an input, we can render frames for arbitrary camera trajectories at test time, including trajectories controlled by a user in the loop. We have built a HTML interface that allows the user to steer our auto-pilot algorithm as it flies through this imaginary world. This demo runs over the internet and is capable of generating a few frames per second. Please see the supplementary video for a demonstration.

\section{Discussion}
\vspace{-0.2em}
We introduce the new problem of perpetual view generation and present a novel framework that combines both geometric and generative techniques as a first step in tackling it. Our system can generate video sequences spanning hundreds of frames, which to our knowledge has not been shown for prior video or view synthesis methods. The results indicate that our hybrid approach is a promising step. Nevertheless, many challenges remain.

First, our render-refine-repeat loop is by design memory-less, an intentional choice which allows us to train on finite length videos yet generate arbitrarily long output using a finite memory and compute budget. As a consequence it aims for local consistency between nearby frames, but does not directly tackle questions of long-term consistency or a global representation.
How to incorporate long-term memory in such a system is an exciting question for future work.
Second, our refinement network, like other GANs, can produce images that seem realistic but not recognizable \cite{hertzmann2019visual}. Further advancements in image and video synthesis generation methods that incorporate geometry would be an interesting future direction. 
Last, we do not model dynamic scenes:
combining our geometry-aware approach with methods that can reason about object dynamics could be another fruitful direction.

\vfill\hrule\medskip\noindent{\small\textbf{Acknowledgements.}
We would like to thank Dominik Kaeser for directing and helping prepare our videos and Huiwen Chang for making the MiDaS models easily accessible. 
}

\clearpage
{\small
\bibliographystyle{ieee_fullname}
\bibliography{reference}

\begin{thebibliography}{10}\itemsep=-1pt

\bibitem{Barnes:2009:PAR}
Connelly Barnes, Eli Shechtman, Adam Finkelstein, and Dan~B Goldman.
\newblock {PatchMatch}: A randomized correspondence algorithm for structural
  image editing.
\newblock {\em ACM Transactions on Graphics (Proc. SIGGRAPH)}, 28(3), Aug.
  2009.

\bibitem{blank2005actions}
Moshe Blank, Lena Gorelick, Eli Shechtman, Michal Irani, and Ronen Basri.
\newblock Actions as space-time shapes.
\newblock In {\em ICCV}, pages 1395--1402. IEEE, 2005.

\bibitem{chaurasia:2013:depth}
Gaurav Chaurasia, Sylvain Duch{\^e}ne, Olga Sorkine-Hornung, and George
  Drettakis.
\newblock Depth synthesis and local warps for plausible image-based navigation.
\newblock {\em Trans. on Graphics}, 32:30:1--30:12, 2013.

\bibitem{chen:2019:youtube3d}
Weifeng Chen, Shengyi Qian, and Jia Deng.
\newblock Learning single-image depth from videos using quality assessment
  networks.
\newblock In {\em The IEEE Conference on Computer Vision and Pattern
  Recognition (CVPR)}, June 2019.

\bibitem{chen2019mono}
Xu Chen, Jie Song, and Otmar Hilliges.
\newblock Monocular neural image based rendering with continuous view control.
\newblock In {\em ICCV}, 2019.

\bibitem{chen2019monocular}
Xu Chen, Jie Song, and Otmar Hilliges.
\newblock Monocular neural image based rendering with continuous view control.
\newblock In {\em ICCV}, pages 4090--4100, 2019.

\bibitem{extremeview}
Inchang Choi, Orazio Gallo, Alejandro Troccoli, Min~H Kim, and Jan Kautz.
\newblock Extreme view synthesis.
\newblock In {\em Proceedings of the IEEE International Conference on Computer
  Vision}, pages 7781--7790, 2019.

\bibitem{choi2019extreme}
Inchang Choi, Orazio Gallo, Alejandro Troccoli, Min~H Kim, and Jan Kautz.
\newblock Extreme view synthesis.
\newblock In {\em ICCV}, pages 7781--7790, 2019.

\bibitem{DVD}
Aidan Clark, Jeff Donahue, and Karen Simonyan.
\newblock Efficient video generation on complex datasets.
\newblock {\em arXiv preprint arXiv:1907.06571}, 2019.

\bibitem{denton2018stochastic}
Emily Denton and Rob Fergus.
\newblock Stochastic video generation with a learned prior.
\newblock {\em arXiv preprint arXiv:1802.07687}, 2018.

\bibitem{finn2016unsupervised}
Chelsea Finn, Ian Goodfellow, and Sergey Levine.
\newblock Unsupervised learning for physical interaction through video
  prediction.
\newblock In {\em NeurIPS}, pages 64--72, 2016.

\bibitem{flynn:2019:deepview}
John Flynn, Michael Broxton, Paul Debevec, Matthew DuVall, Graham Fyffe, Ryan
  Overbeck, Noah Snavely, and Richard Tucker.
\newblock Deepview: View synthesis with learned gradient descent.
\newblock In {\em The IEEE Conference on Computer Vision and Pattern
  Recognition (CVPR)}, June 2019.

\bibitem{Genova_2018_CVPR}
Kyle Genova, Forrester Cole, Aaron Maschinot, Aaron Sarna, Daniel Vlasic, and
  William~T. Freeman.
\newblock Unsupervised training for 3d morphable model regression.
\newblock In {\em The IEEE Conference on Computer Vision and Pattern
  Recognition (CVPR)}, June 2018.

\bibitem{glasner2009super}
Daniel Glasner, Shai Bagon, and Michal Irani.
\newblock Super-resolution from a single image.
\newblock In {\em ICCV}, pages 349--356, 2009.

\bibitem{hays2007scene}
James Hays and Alexei~A Efros.
\newblock Scene completion using millions of photographs.
\newblock {\em ACM Transactions on Graphics (TOG)}, 26(3):4--es, 2007.

\bibitem{hertzmann2019visual}
Aaron Hertzmann.
\newblock Visual indeterminacy in generative neural art.
\newblock {\em arXiv preprint arXiv:1910.04639}, 2019.

\bibitem{fid}
Martin Heusel, Hubert Ramsauer, Thomas Unterthiner, Bernhard Nessler, and Sepp
  Hochreiter.
\newblock Gans trained by a two time-scale update rule converge to a local nash
  equilibrium.
\newblock In {\em NeurIPS}, pages 6626--6637, 2017.

\bibitem{johnson2016perceptual}
Justin Johnson, Alexandre Alahi, and Li Fei-Fei.
\newblock Perceptual losses for real-time style transfer and super-resolution.
\newblock In {\em European conference on computer vision}, pages 694--711.
  Springer, 2016.

\bibitem{Kaneva_2010}
Biliana Kaneva, Josef Sivic, Antonio Torralba, Shai Avidan, and William~T.
  Freeman.
\newblock Infinite images: Creating and exploring a large photorealistic
  virtual space.
\newblock In {\em Proceedings of the IEEE}, 2010.

\bibitem{karras2019style}
Tero Karras, Samuli Laine, and Timo Aila.
\newblock A style-based generator architecture for generative adversarial
  networks.
\newblock In {\em Proceedings of the IEEE Conference on Computer Vision and
  Pattern Recognition}, pages 4401--4410, 2019.

\bibitem{kingma2013auto}
Diederik~P Kingma and Max Welling.
\newblock Auto-encoding variational bayes.
\newblock {\em arXiv preprint arXiv:1312.6114}, 2013.

\bibitem{ledig:2017:photo}
Christian Ledig, Lucas Theis, Ferenc Husz{\'a}r, Jose Caballero, Andrew
  Cunningham, Alejandro Acosta, Andrew~P Aitken, Alykhan Tejani, Johannes Totz,
  Zehan Wang, et~al.
\newblock Photo-realistic single image super-resolution using a generative
  adversarial network.
\newblock In {\em CVPR}, 2017.

\bibitem{levoy:1996:lightfield}
Marc Levoy and Pat Hanrahan.
\newblock Light field rendering.
\newblock In {\em Proceedings of SIGGRAPH 96}, Annual Conference Series, 1996.

\bibitem{mildenhall:2019:llff}
Ben Mildenhall, Pratul~P. Srinivasan, Rodrigo Ortiz-Cayon, Nima~Khademi
  Kalantari, Ravi Ramamoorthi, Ren Ng, and Abhishek Kar.
\newblock Local light field fusion: Practical view synthesis with prescriptive
  sampling guidelines.
\newblock {\em ACM Transactions on Graphics (TOG)}, 2019.

\bibitem{niklaus:2019:kenburns}
Simon Niklaus, Long Mai, Jimei Yang, and Feng Liu.
\newblock 3{D} {Ken} {Burns} effect from a single image.
\newblock {\em ACM Transactions on Graphics (TOG)}, 2019.

\bibitem{park2019SPADE}
Taesung Park, Ming-Yu Liu, Ting-Chun Wang, and Jun-Yan Zhu.
\newblock Semantic image synthesis with spatially-adaptive normalization.
\newblock In {\em Proceedings of the IEEE Conference on Computer Vision and
  Pattern Recognition}, 2019.

\bibitem{ranftl:2020:towards}
Ren{\'e} Ranftl, Katrin Lasinger, David Hafner, Konrad Schindler, and Vladlen
  Koltun.
\newblock Towards robust monocular depth estimation: Mixing datasets for
  zero-shot cross-dataset transfer.
\newblock {\em IEEE Transactions on Pattern Analysis and Machine Intelligence},
  2020.

\bibitem{ross2011reduction}
St{\'e}phane Ross, Geoffrey Gordon, and Drew Bagnell.
\newblock A reduction of imitation learning and structured prediction to
  no-regret online learning.
\newblock In {\em Proceedings of the fourteenth international conference on
  artificial intelligence and statistics}, pages 627--635, 2011.

\bibitem{schodl2000video}
Arno Sch{\"o}dl, Richard Szeliski, David~H Salesin, and Irfan Essa.
\newblock Video textures.
\newblock In {\em Proceedings of the 27th annual conference on Computer
  graphics and interactive techniques}, pages 489--498, 2000.

\bibitem{shaham2019singan}
Tamar~Rott Shaham, Tali Dekel, and Tomer Michaeli.
\newblock Singan: Learning a generative model from a single natural image.
\newblock In {\em ICCV}, pages 4570--4580, 2019.

\bibitem{shi:2014:lightfield}
Lixin Shi, Haitham Hassanieh, Abe Davis, Dina Katabi, and Fredo Durand.
\newblock Light field reconstruction using sparsity in the continuous fourier
  domain.
\newblock {\em Trans. on Graphics}, 34(1):12:1--12:13, Dec. 2014.

\bibitem{Shih3DP20}
Meng-Li Shih, Shih-Yang Su, Johannes Kopf, and Jia-Bin Huang.
\newblock 3d photography using context-aware layered depth inpainting.
\newblock In {\em IEEE Conference on Computer Vision and Pattern Recognition
  (CVPR)}, 2020.

\bibitem{shocher2018ingan}
Assaf Shocher, Shai Bagon, Phillip Isola, and Michal Irani.
\newblock Ingan: Capturing and remapping the {``DNA''} of a natural image.
\newblock {\em arXiv preprint arXiv:1812.00231}, 2018.

\bibitem{shocher2018zero}
Assaf Shocher, Nadav Cohen, and Michal Irani.
\newblock ``zero-shot'' super-resolution using deep internal learning.
\newblock In {\em CVPR}, pages 3118--3126, 2018.

\bibitem{srinivasan:2019:boundaries}
Pratul~P. Srinivasan, Richard Tucker, Jonathan~T. Barron, Ravi Ramamoorthi, Ren
  Ng, and Noah Snavely.
\newblock Pushing the boundaries of view extrapolation with multiplane images.
\newblock In {\em The IEEE Conference on Computer Vision and Pattern
  Recognition (CVPR)}, June 2019.

\bibitem{teterwak2019boundless}
Piotr Teterwak, Aaron Sarna, Dilip Krishnan, Aaron Maschinot, David Belanger,
  Ce Liu, and William~T Freeman.
\newblock Boundless: Generative adversarial networks for image extension.
\newblock In {\em Proceedings of the IEEE International Conference on Computer
  Vision}, pages 10521--10530, 2019.

\bibitem{grf2020}
Alex Trevithick and Bo Yang.
\newblock Grf: Learning a general radiance field for 3d scene representation
  and rendering.
\newblock In {\em arXiv:2010.04595}, 2020.

\bibitem{single_view_mpi}
Richard Tucker and Noah Snavely.
\newblock Single-view view synthesis with multiplane images.
\newblock In {\em The IEEE Conference on Computer Vision and Pattern
  Recognition (CVPR)}, June 2020.

\bibitem{tulsiani:2018:lsi}
Shubham Tulsiani, Richard Tucker, and Noah Snavely.
\newblock Layer-structured 3{D} scene inference via view synthesis.
\newblock In {\em The European Conference on Computer Vision (ECCV)}, September
  2018.

\bibitem{tulyakov2018mocogan}
Sergey Tulyakov, Ming-Yu Liu, Xiaodong Yang, and Jan Kautz.
\newblock Mocogan: Decomposing motion and content for video generation.
\newblock In {\em CVPR}, pages 1526--1535, 2018.

\bibitem{villegas2019high}
Ruben Villegas, Arkanath Pathak, Harini Kannan, Dumitru Erhan, Quoc~V Le, and
  Honglak Lee.
\newblock High fidelity video prediction with large stochastic recurrent neural
  networks.
\newblock In {\em Advances in Neural Information Processing Systems}, pages
  81--91, 2019.

\bibitem{vondrick2016generating}
Carl Vondrick, Hamed Pirsiavash, and Antonio Torralba.
\newblock Generating videos with scene dynamics.
\newblock In {\em NeurIPS}, pages 613--621, 2016.

\bibitem{vondrick2017generating}
Carl Vondrick and Antonio Torralba.
\newblock Generating the future with adversarial transformers.
\newblock In {\em CVPR}, pages 1020--1028, 2017.

\bibitem{wang2019wide}
Yi Wang, Xin Tao, Xiaoyong Shen, and Jiaya Jia.
\newblock Wide-context semantic image extrapolation.
\newblock In {\em Proceedings of the IEEE Conference on Computer Vision and
  Pattern Recognition}, pages 1399--1408, 2019.

\bibitem{wiles2020synsin}
Olivia Wiles, Georgia Gkioxari, Richard Szeliski, and Justin Johnson.
\newblock {SynSin}: {E}nd-to-end view synthesis from a single image.
\newblock In {\em CVPR}, 2020.

\bibitem{yang2019very}
Zongxin Yang, Jian Dong, Ping Liu, Yi Yang, and Shuicheng Yan.
\newblock Very long natural scenery image prediction by outpainting.
\newblock In {\em Proceedings of the IEEE International Conference on Computer
  Vision}, pages 10561--10570, 2019.

\bibitem{Ye_2019_ICCV}
Yufei Ye, Maneesh Singh, Abhinav Gupta, and Shubham Tulsiani.
\newblock Compositional video prediction.
\newblock In {\em ICCV}, 2019.

\bibitem{yu2020pixelnerf}
Alex Yu, Vickie Ye, Matthew Tancik, and Angjoo Kanazawa.
\newblock {pixelNeRF}: {N}eural radiance fields from one or few images.
\newblock In {\em CVPR}, 2021.

\bibitem{yu2018generative}
Jiahui Yu, Zhe Lin, Jimei Yang, Xiaohui Shen, Xin Lu, and Thomas~S Huang.
\newblock Generative image inpainting with contextual attention.
\newblock {\em arXiv preprint arXiv:1801.07892}, 2018.

\bibitem{yu2018free}
Jiahui Yu, Zhe Lin, Jimei Yang, Xiaohui Shen, Xin Lu, and Thomas~S Huang.
\newblock Free-form image inpainting with gated convolution.
\newblock In {\em ICCV}, 2019.

\bibitem{zhang:2018:lpips}
Richard Zhang, Phillip Isola, Alexei~A. Efros, Eli Shechtman, and Oliver Wang.
\newblock The unreasonable effectiveness of deep features as a perceptual
  metric.
\newblock In {\em CVPR}, 2018.

\bibitem{zhou:2018:stereo}
Tinghui Zhou, Richard Tucker, John Flynn, Graham Fyffe, and Noah Snavely.
\newblock Stereo magnification: Learning view synthesis using multiplane
  images.
\newblock {\em ACM Trans. Graph.}, 37(4):65:1--65:12, 2018.

\bibitem{zhou2018non}
Yang Zhou, Zhen Zhu, Xiang Bai, Dani Lischinski, Daniel Cohen-Or, and Hui
  Huang.
\newblock Non-stationary texture synthesis by adversarial expansion.
\newblock {\em arXiv preprint arXiv:1805.04487}, 2018.

\end{thebibliography}
}

\clearpage
\appendix
\twocolumn[{%
	\begin{center}
		\newcommand{\teaserwidth}{\textwidth}
		\vspace{-3mm}
		\centerline{
			  \includegraphics[width=\textwidth]{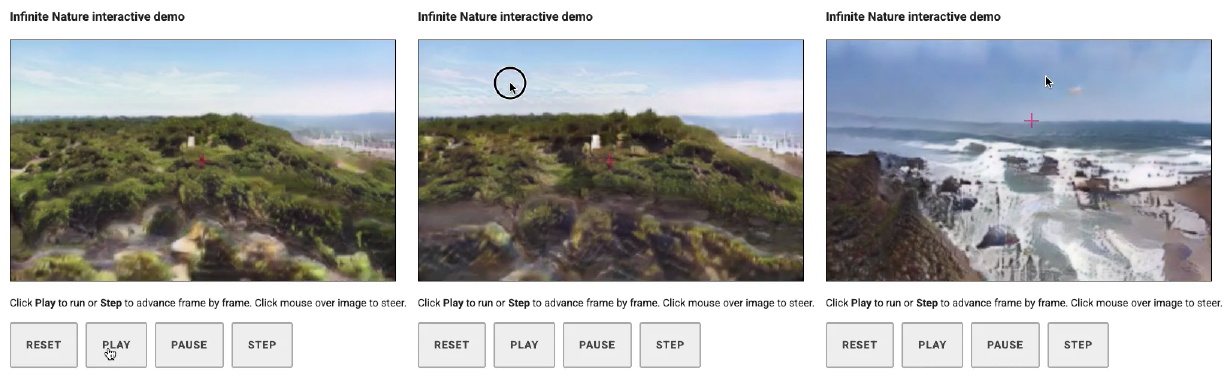}
			   }
			   \captionof{figure}{\textbf{Infinite Nature Demo.} We built a lightweight demo interface so a user can run Infinite Nature and control the camera trajectory. In addition, the demo can take any uploaded image, and the system will automatically run MiDaS to generate an initial depth map, then allow the user hit ``play'' to navigate through the generated world and click to turn the camera towards the cursor. The demo runs at several frames per second using a free Google Colab GPU-enabled backend. Please see our video for the full effect of generating an interactive scene flythrough.
			   }
			   \vspace{-2mm}
			   \label{fig:demo}

	\end{center}%
	}]

\section*{Appendix}

	\section{Implementation Details}
	This section contains additional implementation details for our system, including data generation, network architecture, and inference procedure.

	\subsection{ACID Collection and Processing}
	To create the ACID dataset, we began by identifying over 150
	proper nouns related to coastline and island locations such as \textit{Big Sur}, \textit{Half Moon Bay}, \textit{Moloka'i}, \textit{Shi Shi Beach}, \textit{Waimea bay}, etc. 
	We combined each proper noun with a set of keywords ($\{\textit{aerial}, \textit{drone}, \textit{dji}, and \textit{mavic}\}$) and used these combinations of keywords to perform YouTube video search queries.
	We combined the top 10 video IDs from each query to form a set of candidate videos for our dataset.

	We process all the videos through a SLAM and SfM pipeline as in Zhou \etal~\cite{zhou:2018:stereo}. For each video, this process yields a set of camera trajectories, each containing camera poses corresponding to individual video frames. The pipeline also produces a set of 3D keypoints. We manually identify and remove videos that feature a static camera or are not aerial, as well as videos that feature a large number of people or man-made structures. In an effort to limit the potential privacy concerns of our work, we also discard frames that feature people. In particular, we run the state of the art object detection network~\cite{tan2020efficientdet} to identify any humans present in the frames. If detected humans occupy more than 10\% of a given frame, we discard that frame. The above filtering steps are applied in order to identify high-quality video sequences for training with limited privacy implications, and the remaining videos form our dataset.

	Many videos, especially those that feature drone footage, are shot with cinematic horizontal borders, achieving a letterbox effect. We pre-process every frame to remove detected letterboxes and adjust the camera intrinsics accordingly to reflect this crop operation.

	For the remaining sequences, we run the MiDaS system~\cite{ranftl:2020:towards} on every frame to estimate dense disparity (inverse depth). MiDaS predicts disparity only up to an unknown scale and shift, so for each frame we use the 3D keypoints produced by running SfM to compute scale and shift parameters that best fit the MiDaS disparity values to the 3D keypoints visible in that frame. This results in disparity images that better align with the SfM camera trajectories during training. More specifically, the scale $a$ and shift $b$ are calculated via least-squares as:
	\begin{equation}
		   \argmin_{a, b} \sum_{(x, y, z) \in \mathcal{K}} \big(a\hat{D}_{xyz} + b - z^{-1}\big)^2 
	\end{equation}
	where $\mathcal{K}$ is the set of visible 3D keypoints from the local frame's camera viewpoint, $\hat{D}$ is the disparity map predicted by MiDaS for that frame, and $\hat{D}_{xyz}$ is the disparity value sampled from that map at texture coordinates corresponding to the projection of the point $(x, y, z)$ according to the camera intrinsics. The disparity map $D$ we use during training and rendering is then $D = a\hat{D} + b$.

	\begin{figure}
		\includegraphics[width=\linewidth]{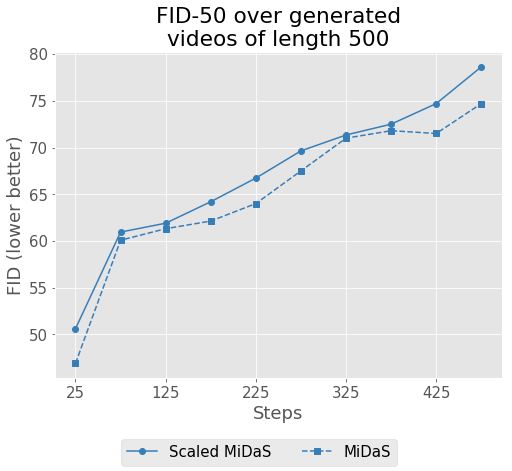}
		\caption{\textbf{Scaled MiDaS vs original MiDaS.} We scale the MiDaS disparity maps to be consistent with the camera poses estimated by SfM during training. At test-time our approach only requires a single image with disparity. Here we show results of 
		FID-50 long generation using the original MiDaS output vs the scaled MiDaS. Despite being only trained on scaled disparity, our model still performs competitively with (unscaled) MiDaS as its input.} 
		\label{fig:midas_plot}
	\end{figure}

	\subsection{Inference without Disparity Scaling}
	Scaling and shifting the disparity as described above requires a sparse point cloud, which is generated from SfM and in turn requires video or multi-view imagery. At test-time, however, we assume only a single view is available. Fortunately, this is not a problem in practice, as scaling and shifting the disparity is only necessary if we seek to compare generated frames at target poses against ground truth. If we just want to generate sequences, we can equally well use the original MiDaS disparity predictions.
	\citefig{fig:midas_plot} compares long generation using scaled and original MiDaS outputs, and shows that using original MiDaS outputs has a negligible effect on the FID scores. \citefig{fig:single_image} shows an example of a long sequence generated with the unscaled MiDaS prediction from a photo taken on a smartphone, demonstrating that our framework runs well on a single test image using the original MiDaS disparity.

	\begin{figure*}[h]
		  \centering
		    \includegraphics[width=\textwidth]{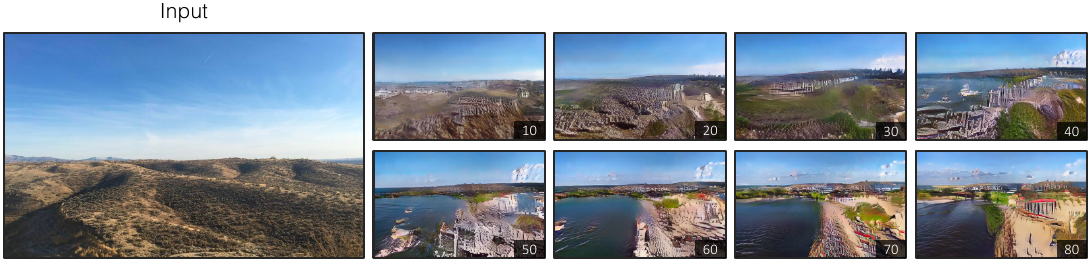}
		      \caption{ {\bf Generation from smartphone photo.} Our perpetual view generation applied to a photo captured by the authors on a smartphone. We use MiDaS for the initial disparity, and assume a field of view of $90^\circ$.}
		        \label{fig:single_image}
	\end{figure*}

	\subsection{Aligning Camera Speed}
	\label{ssec:align_cam}
	The speed of camera motion varies widely in our collected videos, so we normalize the amount of motion present in training image sequences by computing a proxy for camera speed.
	We use the translation magnitude of the estimated camera poses between frames after scale-normalizing the video as in Zhou \etal~\cite{zhou:2018:stereo} to determine a range of rates at which each sequence can be subsampled in order to obtain a camera speed within a desired target range. We randomly select frame rates within this range to subsample videos.
	We picked a target speed range for training sequences that varies by up to 30\% and, on average, leaves 90\% of an image's content visible in the next sampled frame.

	\subsection{Network Architecture}
	We use Spatially Adaptive Normalization (SPADE) of Park \etal~\cite{park2019SPADE} as the basis for our refinement network. The generator consists of two parts, a variational image encoder and a SPADE generator. The variational image encoder maps a given image to the parameters of a multivariate Gaussian that represents its feature. We can use this new distribution to sample GAN noise used by the SPADE generator. We use the initial RGBD frame of a sequence as input to the encoder to obtain this distribution before repeatedly sampling from it (or using its mean at test-time) at every step of refinement. 

	Our SPADE generator is identical to the original SPADE architecture, except that the input has only 5 channels corresponding to RGB texture, disparity, and a mask channel indicating missing regions. 

	We also considered a U-net~\cite{unet}--based approach by using the generator implementation of Pix2Pix~\cite{pix2pix2017}, but found that such an approach struggles to achieve good results, taking longer to converge and in many cases, completely failing when evaluating beyond the initial five steps.

	As our discriminator, we use the Pix2PixHD~\cite{wang2018pix2pixHD} multi-scale discriminator with two scales over generated RGBD frames. To make efficient use of memory, we run the discriminator on random crops of pixels and random generated frames over time.

	\subsection{Loss Weights}
	We used a subset of our training set to sweep over checkpoints and hyperparameter configurations. For our loss, we used $\lambda_{\text{reconst}} = 2$, $\lambda_{\text{perceptual}} = 0.01$, $\lambda_{\text{adversarial}}=1$, $\lambda_{\text{KLD}}=0.05$, $\lambda_{\text{feat matching}}=10$.

	\subsection{Data Source for Qualitative Illustrations}
	Note that for license reasons, we do not show generated qualitative figures and results on ACID. Instead, we collect input images with open source licenses from Pexels~\cite{pexel} and show the corresponding qualitative results in the paper and the supplementary video. The quantitative results are computed on the ACID test set.

		\subsection{Auto-pilot View Control}
		\label{ssec:autocruise}
		We use an auto-pilot 
		view control algorithm when generating long sequences from a single input RGB-D image. This algorithm must generate the camera trajectory in tandem with the image generation, so that it can avoid crashing into the 
		ground or obstacles in the scene. 
		Our basic approach works as follows: at each step we take the current disparity image and categorize all points with disparity below a certain threshold as \textit{sky} and all points with disparity above a second, higher threshold as \textit{near}. (In our experiments these thresholds are set to 0.05 and 0.5.) Then we apply three simple heuristics for view-control: (1) look up or down so that a given percentage (typically 30\%) of the image is \textit{sky}, (2) look left or right, towards whichever side has more \textit{sky}, (3) If more than 20\% of the image is \textit{near}, move up (and if less, down), otherwise move towards a horizontally-centered point 30\% of the way from the top of the image.
		These heuristics determine a (camera-relative) \textit{target look direction} and \textit{target movement direction}. To ensure smooth camera movement, we interpolate the actual look and movement directions only a small fraction (0.05) of the way to the target directions at each frame. The next camera pose is then produced by moving a set distance in the move direction while looking in the look direction. To generate a wider variety of camera trajectories (as for example in~\citesec{sec:diverse}), or to allow user control, we can add an offset to the target look direction that varies over time: a horizontal sinusoidal variation in the look direction, for example, generates a meandering trajectory.

		This approach generates somewhat reasonable trajectories, but an exciting future direction would be to train a model that learns how to choose each successive camera pose, using the camera poses in our training data.

		We use this auto-pilot algorithm to seamlessly integrate user control and obstacle avoidance in our demo interface which can be seen in \citefig{fig:demo}.

		\begin{figure*}[t]
			  \centering
			    \includegraphics[width=\textwidth,trim=.1cm 0cm 0cm -1cm]{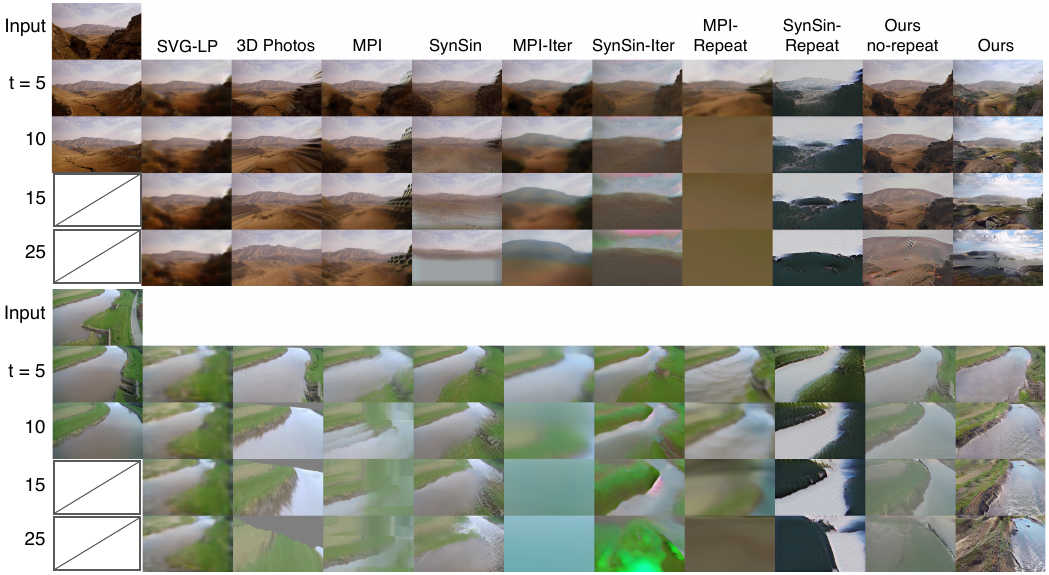}
			      \vspace{-.5em}
			        \caption{ {\bf Additional Qualitative Comparisons.} As in Figure 6 in the main paper, we show more qualitative view synthesis results on various baselines. Notice how other methods produce artifacts like stretched pixels (3D Photos, MPI), or incomplete outpainting (3D Photos, SynSin, Ours no-repeat) or fail to completely move the camera (SVG-LP). Further iter and repeat variants do not improve results. Our approach generates realistic looking images of zoomed in views that involves adding content and super resolving stretched pixels.}
				  \label{fig:rgb_over_time}
				    \vspace{3em}
		\end{figure*}

		\begin{figure*}[h]
			  \centering
			    \includegraphics[width=\textwidth]{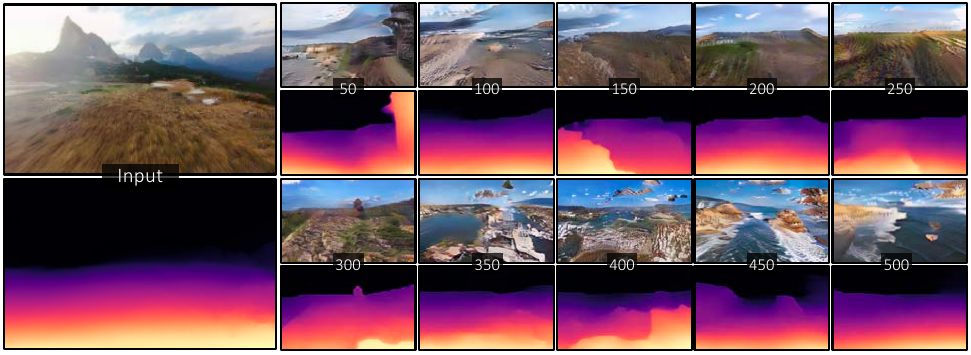}
			      \vspace{-.5em}
			        \caption{ {\bf Long Generation with Disparity.} We show generation of a long sequence with its corresponding disparity output. Our render-refine-repeat approach enables refinement of both geometry and RGB textures.}
				  \label{fig:disparity_over_time}
		\end{figure*}

		\subsection{Additional Frame Interpolation}
		For the purposes of presenting a very smooth and cinematic video with a high frame rate, we can additionally interpolate between frames generated by our model. Since our system produces not just RGB images but also disparity, and since we have camera poses for each frame, we can use this information to aid the interpolation. For each pair of frames $(P_t, I_t, D_t)$ and $(P_{t+1}, I_{t+1}, D_{t+1})$ we proceed as follows:

		First, we create additional camera poses (as many as desired) by linearly interpolating position and look-direction between $P_t$ and $P_{t+1}$. Then, for each new pose $P$ a fraction $\lambda$ of the way between $P_t$ and $P_{t+1}$, we use the differentiable renderer $\mathcal{R}$ to rerender $I_t$ and $I_{t+1}$ from that viewpoint, and blend between the two resulting images:
		\begin{equation}
			\begin{aligned}
				I'_t &= \mathcal{R}(I_t, D_t, P_t, P), \\
				I'_{t+1} &= \mathcal{R}(I_{t+1}, D_{t+1}, P_{t+1}, P), \\
				I &= (1-\lambda) I'_t + \lambda I'_{t+1},
			\end{aligned}
		\end{equation}
		Note: we apply this interpolation to the long trajectory sequences in the supplementary video only, adding four new frames between each pair in the sequence. However, all short-to-mid range comparisons and all figures and metrics in the paper are computed on raw outputs without any interpolation.

		\subsection{Aerial Coastline Imagery Dataset}
		Our ACID dataset is available from our project page at
		\href{https://infinite-nature.github.io/}{https://infinite-nature.github.io}, in the same format as RealEstate10K\cite{zhou:2018:stereo}). For each video we identified as aerial footage of nature scenes, we identified multiple frames for which we compute structure-from-motion poses and intrinsics within a globally consistent system. We divide ACID into train and test splits.

		To get test sequences used during evaluation, we apply the same motion-based frame subsampling described in \citesec{ssec:align_cam} to match the distribution seen during training for all view synthesis approaches. Further we constrain test items to only include forward motion which is defined as trajectories that stay within a $90^\circ$ frontal cone of the first frame. This was done to establish a fair setting with existing view synthesis methods which do not incorporate generative aspects. These same test items were used in the 50-frame FID experiments by repeatedly extrapolating the last two known poses to generate new poses. For the 500-generation FID, we compute future poses using the auto-pilot control described in \citesec{ssec:autocruise}. To get ``real" inception statistics to compare with, we use images from ACID.

		\section{Experimental implementation}

		\subsection{SynSin training}
		We first trained Synsin~\cite{wiles2020synsin} on our nature dataset with the default training settings (i.e. the presets used for the KITTI model). 
		We then modified 
		the default settings by changing the camera stride in order to train Synsin to perform better for the task of longer-range view synthesis.  %
		Specifically, we employ the same motion-based sampling for selecting pairs of images as described in \citesec{ssec:align_cam}. However, here we increase the upper end of the desired motion range by a factor of 5, which allow the network to train with longer camera strides. %
		This obtains a better performance than the default setting, and we use this model for all Synsin evaluations.
		We found no improvement going beyond 5$\times$ camera motion range.  
		We also implemented an exhaustive search for desirable image pairs within a sequence to maximize the training data.

		We also experimented with \textit{SynSin-iter} to synthesize long videos by applying the aforementioned trained SynSin in an auto-regressive fashion at test time. But this performed worse than the direct long-range synthesis.

		In addition to this, we also consider the repeat variant. \textit{SynSin-repeat} was implemented using a similar training setup, however instead we also train SynSin to take its own output and produce the next view for $T=5$ steps. Due to memory and engineering constraints, we are unable to fit \textit{SynSin-repeat} with the original parameters into memory, so we did our best by by reducing the batch size while keeping as faithful to the original implementation. While this does not indicate SynSin fails at perpetual view generation, it does suggest that certain approaches are better suited to solve this problem.

		\begin{figure}[t]
			\includegraphics[width=\linewidth]{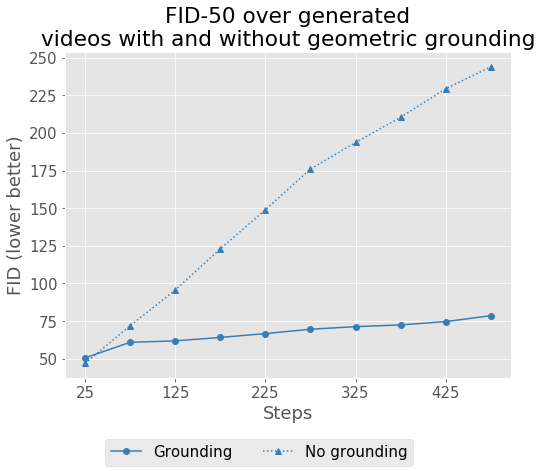}
			\caption{\textbf{Geometric Grounding Ablation.} Geometric grounding is used to explicitly ensure disparities produced by the refinement network match the geometry given by its input. We find this important as otherwise subtle drift can cause the generated results to diverge quickly as visible in \citefig{fig:drift_gg}.} 
			\label{fig:grounding_ablation}
		\end{figure}

		\begin{figure*}[h]
			  \centering
			    \includegraphics[width=\textwidth]{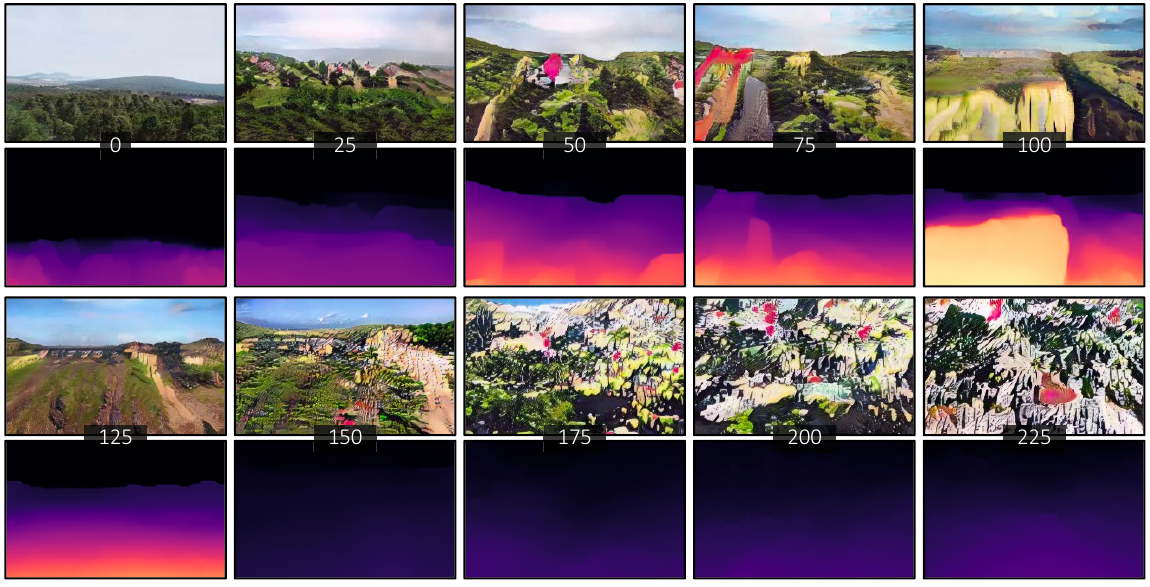}
			      \vspace{-.5em}
			        \caption{ {\bf Geometric Grounding Ablation.} An example of running our pretrained model on the task of long trajectory generation but \emph{without} using geometric grounding. Disparity maps are shown using an \textit{unnormalized} color scale. Athough the output begins plausibly, by the 150th frame the disparity map has drifted very far away, and subsequently the RGB output drifts after the 175th frame.}
				  \label{fig:drift_gg}
		\end{figure*}

		\section{Additional Analysis and Results}
		This section contains additional results and analysis to better understand Infinite Nature's behavior. In \citefig{fig:rgb_over_time}, we show additional view synthesis results given an input image across various baselines.

		\subsection{Limitations}
		As discussed in the main paper, our approach is essentially a memory-less Markov process that does not guarantee global consistency across multiple iterations. This manifests in two ways: First on the geometry, \ie when you look back, there is no guarantee that the same geometric structure that was observed in the past will be there. Second, there is also no global consistency enforced on the appearance—--the appearance of the scene may change in short range, such as sunny coastline turning into a cloudy coastline after several iterations. Similarly, after hundreds of steps, two different input images may end up in a scene that has similar stylistic appearance, although never exactly the same set of frames. Adding global memory to a system like ours and ensuring more control over what will happen in the long range synthesis is an exciting future direction.

		\begin{figure*}[p]
			  \centering
			    \includegraphics[width=\linewidth]{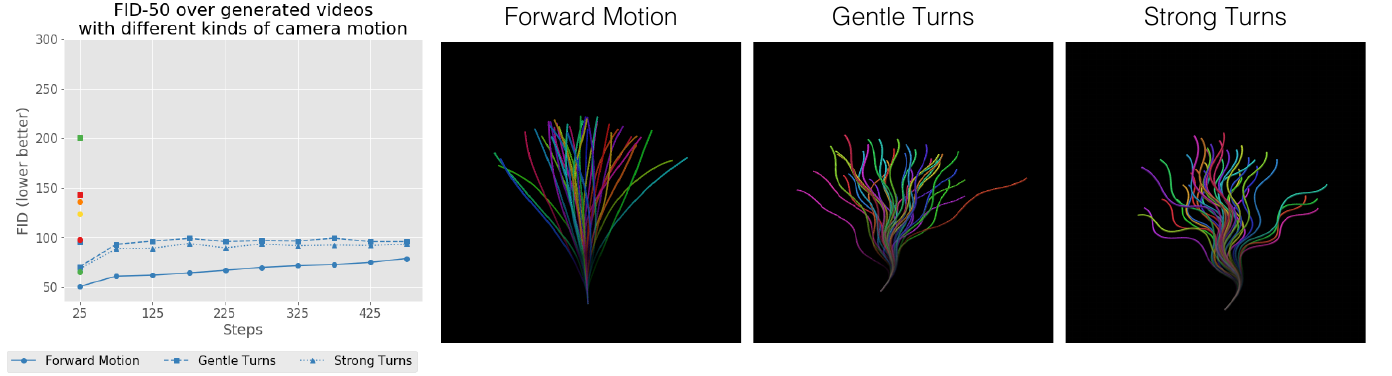}
			      \vspace{-.5em}
			        \caption{ {\bf FID with different camera motion.} We consider different types of camera motion generated by our auto-pilot algorithm with different parameters and its effect on generated quality. \textbf{Right:} Top-down view of three variations of camera motion that add different amounts of additional turning to the auto-pilot algorithm. \textbf{Left:} Even with strongly turning camera motion, our auto-pilot algorithm is able to generate sequences whose quality is only slightly worse than our full model evaluated only on forward translations. The unlabeled points refer to reported baselines on FID-50 from the main paper. See \citesec{sec:diverse}.
				  }
				    \label{fig:different_motion}
		\end{figure*}

		\subsection{Disparity Map}
		In addition to showing the RGB texture, we can also visualize the refined disparity to show the geometry. In \citefig{fig:disparity_over_time}, we show the long generation as well as its visualized disparity map in an unnormalized color scheme. Note that the disparity maps look plausible as well because we train our discriminator over RGB and disparity concatenated. Please also see our results in the supplementary video.

		\subsection{Effect of Disabling Geometric Grounding}
		We use geometric grounding as a technique to avoid drift. In particular we found that without this grounding, over a time period of many frames the render-refine-repeat loop gradually pushes disparity to very small (i.e.\ distant) values. \citefig{fig:drift_gg} shows an example of this drifting disparity: the sequence begins plausibly but before frame 150 is reached, the disparity (here shown unnormalized) has become very small. It is notable that once this happens the RGB images then begin to deteriorate, drifting further away from the space of plausible scenes. Note that this is a test-time difference only: the results in \citefig{fig:drift_gg} were generated using the same model checkpoint as our other results, but with geometric grounding disabled at test time. We show FID-50 results to quantitatively measure the impact of drifting in \citefig{fig:grounding_ablation}.

		\subsection{Results under Various Camera Motions}
		\label{sec:diverse}
		In addition to the demo, we also provide a quantitative experiment to measure how the model's quality changes with different kinds of camera motion over long trajectories.
		As described in \citesec{ssec:autocruise}, our auto-pilot algorithm can be steered by adding an offset to the target look direction. We add a horizontal offset which varies sinusoidally, causing the camera to turn alternately left and right every 50 frames. \citefig{fig:different_motion} compares the FID-50 scores of sequences generated where the relative magnitude of this offset is 0.0 (no offset), 0.5 (gentle turns), and 1.0 (stronger turns), and visualizes the resulting camera trajectories, viewed from above. This experiment shows that our method is resilient to different turning camera motions, with FID-50 scores that are comparable on long generation.

		\subsection{Generating Forward-Backwards Sequences}
		\label{sec:palindrome}
		Because the \textit{Render-Refine-Repeat} framework uses a memory-less representation to generate sequences, the appearance of content is not maintained across iterations. As a consequence, pixel content seen in one view is not guaranteed to be preserved later when seen again, particularly if it goes out of frame. We can observe such inconsistency by synthesizing forward camera motion followed by the same motion backwards (a palindromic camera trajectory), ending at the initial pose. While generating the forward sequence of frames, some of the content in the original input image will leave the field of view. Then, when synthesizing the backward motion, the model must regenerate this forgotten content anew, resulting in pixels that do not match the original input. \citefig{fig:palindrome} shows various input scenes generated for different lengths of forward-backward motion. The further the camera moves before returning to the initial position, the more content will leave the field of view, and so we find that that longer the palindromic sequence, the more the image generated upon returning to the initial pose will differ from the original input image.
		\begin{figure*}[p]
			  \centering
			  \includegraphics[width=\linewidth]{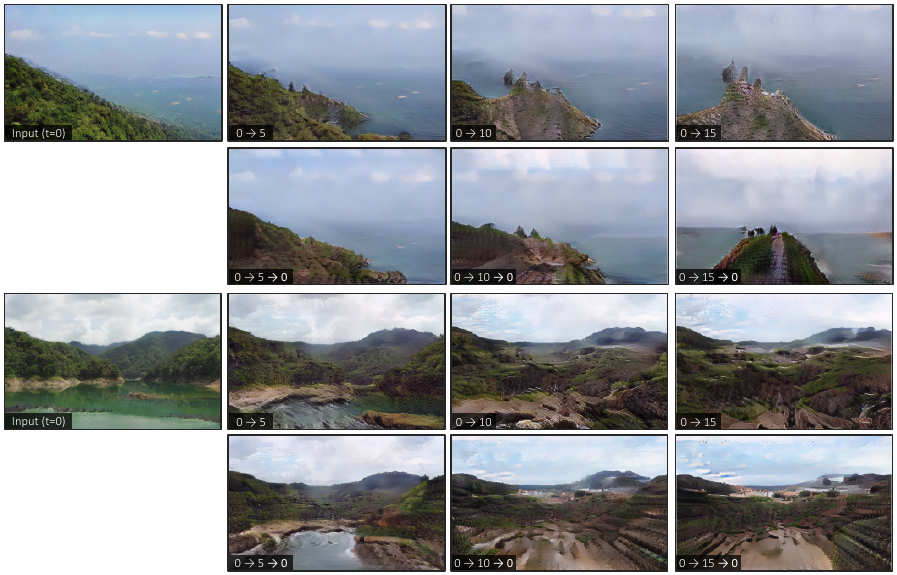}
			  \vspace{-.5em}
			  \caption{\textbf{Palindromic Poses.} Here we show Infinite Nature generated on palindromic sequences of poses of different lengths. Because our model uses a memory-less representation, the forward-backward motion requires the model to hallucinate content it has previously seen but which has gone out frame or been occluded, resulting in a generated image that does not match the original input.} 
			  \label{fig:palindrome}
		\end{figure*}

\end{document}